\documentclass[sigconf, nonacm]{acmart}
\AtBeginDocument{%
  }

\usepackage{arydshln}
\usepackage[utf8]{inputenc} 
\usepackage[T1]{fontenc}    
\usepackage{hyperref}       
\usepackage{graphicx}
\usepackage{url}            
\usepackage{booktabs}       
\usepackage{amsfonts}       
\usepackage{nicefrac}       
\usepackage{microtype}      
\usepackage{float}
\usepackage{xcolor}
\usepackage{subcaption}
\usepackage{multirow}
\usepackage{enumitem}
\usepackage{mathrsfs}
\usepackage{wrapfig}

\setcopyright{acmlicensed}
\copyrightyear{2018}
\acmYear{2018}
\acmDOI{XXXXXXX.XXXXXXX}
\acmConference[Conference acronym 'XX]{Make sure to enter the correct
  conference title from your rights confirmation email}{June 03--05,
  2018}{Woodstock, NY}

\begin{document}

\title{Hierarchically Disentangled Recurrent Network for Factorizing System Dynamics of Multi-scale Systems: An application on Hydrological Systems}

\author{Rahul Ghosh}
\email{ghosh128@umn.edu}
\affiliation{%
  \institution{University of Minnesota - Twin Cities}
  \city{Minneapolis}
  \state{Minnesota}
  \country{USA}
}

\author{Arvind Renganathan}
\email{renga@umn.edu}
\affiliation{%
  \institution{University of Minnesota - Twin Cities}
  \city{Minneapolis}
  \state{Minnesota}
  \country{USA}
}

\author{Zachary P. McEachran}
\email{zachary.mceachran@noaa.gov}
\affiliation{%
  \institution{NOAA National Weather Service}
  \city{Chanhassen}
  \state{Minnesota}
  \country{USA}
}

\author{Somya Sharma}
\email{sharm636@umn.edu}
\affiliation{%
  \institution{University of Minnesota - Twin Cities}
  \city{Minneapolis}
  \state{Minnesota}
  \country{USA}
}

\author{Kelly Lindsay}
\email{lind0436@umn.edu}
\affiliation{%
  \institution{University of Minnesota - Twin Cities}
  \city{Minneapolis}
  \state{Minnesota}
  \country{USA}
}

\author{Michael Steinbach}
\email{stei0062@umn.edu}
\affiliation{%
  \institution{University of Minnesota - Twin Cities}
  \city{Minneapolis}
  \state{Minnesota}
  \country{USA}
}

\author{John L. Nieber}
\email{nieber@umn.edu}
\affiliation{%
  \institution{University of Minnesota - Twin Cities}
  \city{Minneapolis}
  \state{Minnesota}
  \country{USA}
}

\author{Christopher Duffy}
\email{cxd11@psu.edu}
\affiliation{%
  \institution{Pennsylvania State University}
  \city{State College}
  \state{Pennsylvania}
  \country{USA}
}

\author{Vipin Kumar}
\email{kumar001@umn.edu}
\affiliation{%
  \institution{University of Minnesota - Twin Cities}
  \city{Minneapolis}
  \state{Minnesota}
  \country{USA}
}

\renewcommand{\shortauthors}{Ghosh et al.}


\begin{abstract}
    We present a framework for modeling multi-scale processes, and study its performance in the context of streamflow forecasting in hydrology. Specifically, we propose a novel hierarchical recurrent neural architecture that factorizes the system dynamics at multiple temporal scales and captures their interactions. This framework consists of an inverse and a forward model. The inverse model is used to empirically resolve the system's temporal modes from data (physical model simulations, observed data, or a combination of them from the past), and these states are then used in the forward model to predict streamflow. Experiments on several catchments from the National Weather Service North Central River Forecast Center show that FHNN outperforms standard baselines, including physics-based models and transformer-based approaches. The model demonstrates particular effectiveness in catchments with low runoff ratios and colder climates. We further validate FHNN on the CAMELS (Catchment Attributes and MEteorology for Large-sample Studies), which is a widely used continental-scale hydrology benchmark dataset, confirming consistent performance improvements for 1-7 day streamflow forecasts across diverse hydrological conditions. Additionally, we show that FHNN can maintain accuracy even with limited training data through effective pre-training strategies and training global models.
\end{abstract}

\maketitle

\section{Introduction}
\label{sec:Introduction}

\begin{figure*}[t!]
    \centering
    \includegraphics[width=0.85\linewidth]{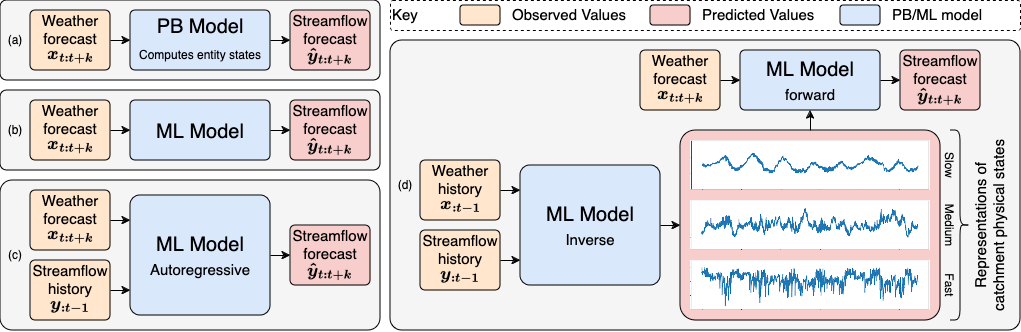}
    \vspace{-0.1in}
    \caption{\footnotesize \textit{Schematic of different modeling approaches for streamflow forecasting illustrating each approach's inputs, processes, and outputs. (a) Process-based Forward Models use weather forecasts to estimate catchment states and generate streamflow forecasts. (b) ML-based Forward Models (e.g., LSTM) directly use weather forecasts to predict streamflow forecasts. (c) ML-based Autoregressive Forward Models utilize weather forecasts and historical streamflow data for streamflow forecasting. (d) The proposed FHNN framework integrates an Inverse Model that generates catchment state representations from historical weather and streamflow data, which are then used alongside weather forecasts in a Forward Model to produce streamflow forecasts, capturing dynamics at multiple temporal scales (slow, medium, fast)}}
    \label{fig:related work}
    \vspace{-0.2in}
\end{figure*}

Physical systems, ranging from natural ecosystems to engineered structures, are characterized by complex interactions of internal states and external drivers. Response of such systems as a function of input drivers is governed by physical states that encapsulate their internal characteristics~\cite{ghosh2023entity}. For instance, in earth sciences, the catchment's streamflow response to meteorological drivers like precipitation and temperature is governed by complex physical processes specific to that catchment (river basin)~\cite{ghosh2022robust}. These processes involve factors such as snow-pack melting and evapotranspiration in soil and plants. Figure~\ref{fig:related work}(a, b, and c) presents schematic representations of Forward models. These forward model provides a structured way to represent these relationships, facilitating improved understanding and prediction of system dynamics in various scientific and engineering contexts.

Over the years, a number of physics-based models (PBMs) have been developed to represent different aspects of physical systems. For example, for streamflow modeling, the hydrological community makes extensive use of rainfall-runoff models~\cite{burnash1973generalizedsacsma} to approximate the underlying bio-geophysical processes of the water cycle, shown in Figure~\ref{fig:related work}(a). Further, these models have been used in operational settings to forecast streamflow and water levels at NOAA’s National Weather Service (NWS). Current operational approaches require a forecaster to assimilate real-time observations into the model, i.e., sequential data assimilation~\cite{bertino2003sequential}.

While PBMs have been the traditional approach for modeling environmental systems, machine learning (ML) models are now increasingly being adopted across scientific domains, including hydrology, where they have outperformed state-of-the-art process-based models in several settings~\cite{feng2020enhancing,willard2022daily}. Figure~\ref{fig:related work} (b) illustrates such ML based workflows. Specifically, Time-aware deep learning techniques such as LSTMs have gained prominence as powerful tools for capturing temporal dependencies and building states~\cite{kratzert2019towards} in environmental systems. While transformers have been shown to be vastly superior to LSTMs in language modeling tasks, they have not been able to outperform LSTMs in environmental problems~\cite{yin2022rr} due to their inability to model states explicitly and the need for large amounts of data (See Section~\ref{sec:Preliminaries and Related Works} for additional discussion). Note that these ML models, when trained in an autoregressive manner~\cite{feng2020enhancing}, can incorporate real time observations into the model's context, to make forecasts into the future in operational settings (as shown in Figure~\ref{fig:related work} (c)). This approach is far more compute efficient than optimization approaches  (e.g., Ensemble Kalman Filtering) that are traditionally used in physical sciences for data assimilation.

The proposed methodology in this paper is meant to enhance LSTMs’ capability to model the underlying states of environmental systems by addressing a crucial gap in the current formulation. Specifically, LSTMs do not represent states explicitly at multiple temporal scales or model their interactions. Environmental systems often feature phenomena that evolve at different time scales, from rapid fluctuations to slowly changing trends. For instance, the impact of the location of rainfall on runoff is evident at hourly/daily and seasonal scales~\cite{zhou2021impacts}. The soil moisture state of a catchment is built up at a weekly scale and thus needs a longer window of information. On the other hand, snow-pack is accumulated at a monthly scale and thus takes longer to impact a catchment's runoff (snow accumulation in December impacts streamflow in April)~\cite{dingman2015physical}. Thus, standard LSTM’s inability to explicitly model such states that are evolving and interacting at different scales, hinders their effectiveness in capturing the full behavior spectrum of complex physical systems with interdependent multi-scale dynamics.


In this paper, we propose a novel factorized hierarchical neural network (FHNN) architecture (Figure~\ref{fig:related work}d) that factorizes the system dynamics at multiple temporal scales and captures their interactions. This framework consists of an inverse and a forward model. The inverse model is used to empirically resolve the system's temporal modes from data (physical model simulations, observed data, or a combination of them from the past), and these states are then used in the forward model to predict streamflow. In a hydrological system, these modes can represent different processes, evolving at different temporal scales (e.g., slow: groundwater recharge vs. fast: surface runoff due to extreme rainfall). 

Two key advantages of our approach are: 1) it can model processes that are evolving and interacting at multiple scales, and 2) once trained, it can incorporate new observations into the model's context (internal state). Experiments with several river catchments from the National Weather Service (NWS) North Central River Forecast Center (NCRFC) region show the efficacy of this ML-based data assimilation framework compared to standard baselines, especially for basins that have a long history of observations. To handle basins with a short observation history, we present two orthogonal strategies for training our FHNN framework: (a) using simulation data from imperfect simulations and (b) using observation data from multiple basins to build a global model.  We show that both of these strategies (that can be used individually or together) are highly effective in mitigating the lack of training data. The improvement in forecast accuracy is particularly noteworthy for basins where local models perform poorly because of data sparsity. While demonstrated in the context of hydrological streamflow modeling, our proposed framework's effectiveness in modeling multi-scale processes makes it useful for a wide variety of science and engineering applications. Our main contributions are listed below:

$\bullet$ We introduce novel FHNN that models multi-scale dynamics by factorizing system states at different temporal scales.

$\bullet$ We demonstrate two effective strategies for managing data sparsity in hydrological modeling: 1) pre-training FHNN on output from a PBM model used by NOAA (SacSMA) and 2) developing a global model trained across multiple basins. Both approaches significantly improve FHNN's performance, given limited observations.

$\bullet$ FHNN demonstrates superior performance over state-of-the-art ML approaches across multiple river basins in the NWS-NCRFC operational region.

$\bullet$ We evaluate FHNN on the CAMELS dataset, a continental-scale hydrology benchmark, showing consistent outperformance over baseline models for 1-day to 7-days streamflow forecasts.

$\bullet$ We provide interpretable visualizations of the learned multi-scale states, offering insights into the model's representation of hydrological processes at different temporal scales.

\section{Problem formulation}
\label{sec:Problem formulation}
This study focuses on learning driver-response behavior for entities. These entities can be physical systems like flux towers, river basins, tasks, people, or domains/distributions. Specifically, we focus on understanding and predicting how precipitation transforms into streamflow at the catchment scale, a key part of operational hydrologic forecasting. This problem is referred to as ``rainfall-runoff transform". For a basin, we have access to multiple driver/response pairs of time series sequences, as $\{(\boldsymbol{x_1}, y_1), (\boldsymbol{x_2}, y_2), \dots, (\boldsymbol{x_T}, y_T)\}$, where, $\boldsymbol{x_t} \in \mathbb{R}^{D_x}$ represents the input vector at time $t \in T$ with $D_x$ dimensions, and $y_t \in \mathbb{R}$ represents the corresponding output. Although in this study, $y_t$ is a single scalar target, it can be multiple targets in many scenarios and thus is a simple extension. The goal is to learn a regression function $\mathcal{F}: X \rightarrow Y$ that maps the input drivers to the output response for an entity.

\section{Preliminaries and Related Works}
\label{sec:Preliminaries and Related Works}

\noindent\textbf{Predictive Modeling for Dynamical Systems:} ML models have been extensively used by the scientific community as emulators of physics-based models~\cite{lall2014debates}. A significant advantage of pure ML is its ability to be entirely data-driven, bypassing the need for causal structures derived from domain knowledge, which can often be incomplete. In hydrology, Long Short-Term Memory (LSTM) networks have been widely used to predict variables like water temperature~\cite{zwart2023near,willard2022daily} and water level~\cite{bowes2019forecasting}, as shown in Figure~\ref{fig:related work} (b). LSTM-based models have achieved state-of-the-art performance in streamflow modeling~\cite{kratzert2019towards} and flood-forecasting~\cite{nearing2024global,yin2021rainfall}, particularly when trained on multi-basin data capture diverse rainfall-runoff responses.

While Transformers have shown superior performance in language modeling and time series forecasting~\cite{tayal2024exotst,YuqietalPatchTST, liu2023itransformer, zhang2023crossformer}, their effectiveness in environmental science is less clear. Unlike LSTMs' sequential processing with evolving internal states, Transformers use self-attention to access all timesteps simultaneously. This distinction matters because environmental systems exhibit cumulative path-dependent effects~\cite{boehrer2008stratification,jia2017state} - for instance, streamflow depends on the sequential evolution of soil moisture and groundwater levels. Transformers also tend to be data hungry. The reason is that the attention mechanisms in transformers are approximating the evolution of states in a path-dependent system through multi-headed attention over all historical timesteps and are essentially memory machines ~\cite{mahdavi2023memorization,geva2020transformer,carlini2022quantifying}. This poses additional challenges given the sparsity of hydrological records. With many regions having limited years of data, LSTM's explicit modeling of sequential state evolution often proves more effective, as confirmed by studies showing Transformer-based models underperform LSTM-based models in hydrological applications~\cite{liu2024probing}.

\noindent\textbf{Forecasting for Dynamical Systems:} The standard LSTM model cannot ingest near-real-time response data and perform data assimilation. Thus, a simple extension to the standard LSTM is to provide lagged response data as input for the model. Several studies~\cite{feng2020enhancing,moshe2020hydronets} have shown that autoregression (AR) improves streamflow predictions from LSTMs. Figure~\ref{fig:related work} (c) shows the diagrammatic representation of the PBM.

\begin{figure*}[t]
    \begin{subfigure}{0.25\linewidth}
        \centering
        \includegraphics[width=\linewidth]{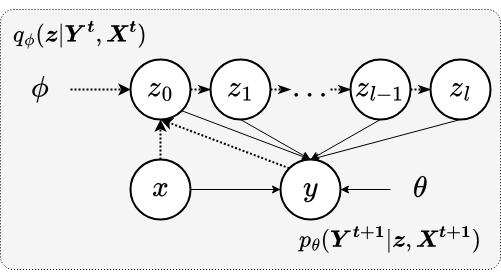}
        \caption{}
        \label{fig:Graph}
    \end{subfigure}
    \begin{subfigure}{0.7\linewidth}
        \centering
        \includegraphics[width=\linewidth]{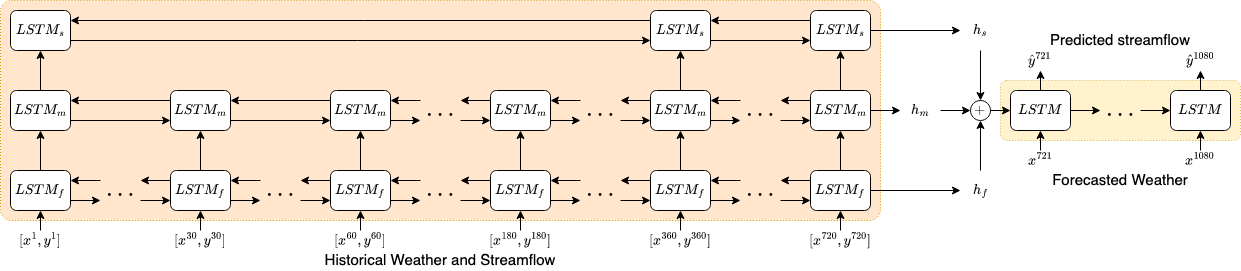}
        \caption{}
        \label{fig:Architecture}
    \end{subfigure}
    \vspace{-0.2in}
    \caption{\footnotesize(a) \textit{Graphical model representation of FHNN for time series data. The model illustrates the relationships between latent variables ($z_0$ to $z_l$), observed variables ($x$ and $y$), and model parameters ($\phi$ and $\theta$). It shows the generative process ($p_{\theta}(\boldsymbol{Y^t}|\boldsymbol{z}, \boldsymbol{X^t})$) and the inference process ($q_{\phi}(\boldsymbol{z}|\boldsymbol{X^{t-1}}, \boldsymbol{Y^{t-1}}$), depicting how past observations influence the current latent state and how the current latent state and inputs generate the output. This structure enables FHNN to capture multi-scale temporal dependencies and generate predictions.}, (b) \textit{Architecture of FHNN for streamflow prediction. The model processes historical weather and streamflow data at three different timescales (fast, medium, and slow) using parallel BiLSTM layers, capturing temporal dependencies at various resolutions. The final hidden states from each timescale ($h_f$, $h_m$, $h_s$) are combined with forecasted weather data to generate streamflow predictions for future time steps. This design allows the model to integrate both short-term fluctuations and long-term trends in the input data for more accurate streamflow forecasting.}}
    \vspace{-0.2in}
\end{figure*}

\noindent\textbf{Knowledge-guided Machine Learning:} Knowledge-guided Machine Learning~\cite{willard2022integrating} demonstrates the benefits of integrating domain knowledge to enhance ML models. One effective approach is the knowledge-guided architecture, which incorporates known hierarchical structures among processes when designing an ML model~\cite{khandelwal2020physics}. These models outperform direct driver-to-response mappings by embedding physical principles into the network architecture and modifying existing components to align with assumptions about interacting physical processes/states. Recently, learnable, differentiable process-based models~\cite{feng2023suitability} with embedded neural networks have been shown to achieve predictive performance comparable to LSTM models. Despite their promise, these models' rigid architectural design poses challenges similar to those of traditional process-based models. Our framework addresses the need to model the internal states of a dynamical system while maintaining the flexibility to allow the model to discover processes not captured in existing domain knowledge.

\section{Method}
\label{sec:method}

Our proposed method infers latent entity states ($\boldsymbol{h} \in \mathbb{R}^{D_h}$) given the historical time-varying driver ($\boldsymbol{X}_{hist} = [\boldsymbol{x^1}, \boldsymbol{x^2}, \dots, \boldsymbol{x^T}]$ where $\boldsymbol{x^t} \in \mathbb{R}^{D_x}$) and response ($\boldsymbol{Y}_{hist} = [y^1, y^2, \dots, y^T]$ where $y^t \in \mathbb{R}$) data. Our approach uses these inferred latent states to forecast an entity's response ($\boldsymbol{Y}_{forecast} = [y^{T+1}, \dots, y^{T+K}]$) given the forecasted drivers ($\boldsymbol{X}_{forecast} = [\boldsymbol{x^{T+1}}, \dots, \boldsymbol{x^{T+K}}]$). However, effective modeling of the internal states of physical entities requires capturing dynamics with varying temporal frequency. We introduce a Factorized Hierarchical Neural Network (FHNN) that enables robust modeling of internal states by capturing and factorizing information at various temporal scales. Unlike standard RNNs, which treat all time steps equally, FHNN extracts dependencies that may be short, medium, or long-term. Specifically, FHNN consists of a sequence encoder (inference network) and a decoder (generator network), as shown in Figure.~\ref{fig:Architecture}. By hierarchically organizing the modeling process in the sequence encoder, FHNN can effectively capture and process information across a wide range of temporal distances, making it well-suited for tasks where understanding complex, interconnected patterns over extended sequences is paramount. The modeled internal states are then used by the generator network to forecast the response using the modeled states and future drivers. In the following sections, we will delve deeper into the architecture and mechanisms of FHNN, illustrating how it overcomes the limitations of standard RNNs and empowers us to tackle challenging problems involving long-range dependencies in sequential data.

\subsection{State Encoder}
The inference network ($q_{\phi}: \mathbb{R}^{T \times (D_x + 1)} \rightarrow \mathbb{R}^{D_z}$) is trained to encode the current state of the entity into the latent space using a hierarchical neural network. Our formulation captures the general intuition that we can separate the entity's state into its dynamic components (described by $z_0, z_1, \dots, z_l$) of different time scales. Our framework can be applied to model any number of such internal states, as shown in Figure~\ref{fig:Graph}. In this study, we model the internal state of a hydrological basin using three latent variables (slow, medium, and fast), i.e., $z_s$, $z_m$, and $z_f$. Thus, our latent state has a factorized form:

{

\begin{equation}
    \begin{split}
        q(\boldsymbol{z}|\boldsymbol{y_{1:t}}, \boldsymbol{x_{1:t}}) &= q(\boldsymbol{z_s},\boldsymbol{z_m},\boldsymbol{z_f}|\boldsymbol{y_{1:t}}, \boldsymbol{x_{1:t}})\\
        &= q(\boldsymbol{z_s},\boldsymbol{z_m}|\boldsymbol{z_f},\boldsymbol{y_{1:t}}, \boldsymbol{x_{1:t}})q(\boldsymbol{z_f}|\boldsymbol{y_{1:t}}, \boldsymbol{x_{1:t}})\\ 
        &= q(\boldsymbol{z_s}|\boldsymbol{z_m},\boldsymbol{z_f})q(\boldsymbol{z_m}|\boldsymbol{z_f})q(\boldsymbol{z_f}|\boldsymbol{y_{1:t}}, \boldsymbol{x_{1:t}})
    \end{split}
\end{equation}
}

Particularly, in this configuration, the encoder consists of three LSTM networks, each processing sequence with skip connections. The lowest LSTM models the data at the finer resolution $\Delta t$, the middle LSTM  at the medium resolution $m\Delta t$, and the topmost LSTM models the data at the coarsest resolution $s\Delta t$. We implement this using bidirectional-LSTM-based sequence encoders. LSTM~\cite{graves2005framewise} is particularly suited for our task where long-range temporal dependencies between driver and response exist as they are designed to avoid exploding and vanishing gradient problems. The final hidden states for the forward ($\boldsymbol{h_f}$) and backward LSTM ($\boldsymbol{h_b}$) are added to get the final embeddings $\boldsymbol{h}$ for each LSTM network, as shown in Figure~\ref{fig:Architecture}. The dynamical system's internal state ($z$) is obtained by first concatenating the slow, medium, and fast states and feeding them into a multi-layer perceptron (MLP). Thus, the sequence encoder functions are of the form:

{

\begin{equation}
    \begin{split}
        [\boldsymbol{h^f_t}]_{_{1:T}}^{^{\Delta T}} &= \text{BiLSTM}([\boldsymbol{x_t};y_t]_{_{1:T}}^{^{\Delta T}};\phi_{\boldsymbol{h_f}})\\
        [\boldsymbol{h^m_t}]_{_{1:T}}^{^{m\Delta T}} &= \text{BiLSTM}([\boldsymbol{h^f_t}]_{_{1:T}}^{^{m\Delta T}};\phi_{\boldsymbol{h_m}})\\
        [\boldsymbol{h^s_t}]_{_{1:T}}^{^{s\Delta T}} &= \text{BiLSTM}([\boldsymbol{h^m_t};\boldsymbol{h^f_t}]_{_{1:T}}^{^{s\Delta T}};\phi_{\boldsymbol{h_s}})\\
        \boldsymbol{z} &= \text{MLP}([\boldsymbol{h^s_T};\boldsymbol{h^m_T};\boldsymbol{h^f_T}];\phi_{\boldsymbol{z}})
    \end{split}
\end{equation}
}

where the encoder $\mathcal{E}(\boldsymbol{y_{1:t}}, \boldsymbol{x_{1:t}};\phi)$ parameter set $\phi$ is divided into the LSTM parameters ($\phi_{\boldsymbol{h}}$) and the MLP parameters ($\phi_{\boldsymbol{z}}$). Depending on the task, the encoder can comprise multiple layers, each with several RNNs.

\subsection{Response Decoder}
The generator network allows for the conditional generation of response data given the latent variable ($\boldsymbol{z}$) from the decoder and the driver data. The conditional generative process of the model is given in Figure~\ref{fig:Graph} as follows: for a sequence of historical driver and response data ($\boldsymbol{X}_{hist}$ and $\boldsymbol{Y}_{hist}$), $\boldsymbol{z}$ is obtained from the hierarchical sequence encoder $q_{\phi}(\boldsymbol{z}|[\boldsymbol{X}_{hist};\boldsymbol{Y}_{hist}])$, and the sequence of response data for the forecast time-period is generated from the decoder $p_{\theta}(\boldsymbol{Y}_{forecast}|\boldsymbol{z}, \boldsymbol{X}_{forecast})$. Specifically, we construct an LSTM-based conditional sequence generator $y_t = LSTM(\boldsymbol{z}, \boldsymbol{x_{T+1:t}};\theta)$, where the encoded internal state of the system ($\boldsymbol{z}$) is used as the initial state ($\boldsymbol{h}^0$) of the LSTM model. We denote the decoder based forward model using the symbol $\mathcal{F}(\boldsymbol{z}, \boldsymbol{x_{T+1:t}};\theta)$.

The model is trained in an end-to-end fashion on the training windows from the training data. The final objective of FHNN during training can be formally expressed as:

{
\vspace{-0.2in}
\begin{equation}
    \label{eq:Loss}
    \begin{split}
        \arg \min_{\phi, \theta} &\quad\frac{1}{K}\sum_{t=T+1}^{T+K} (y^t - \mathcal{F}(\boldsymbol{z}, \boldsymbol{x_{T+1:t}};\theta))^2\\
        where &\quad z = \mathcal{E}(\boldsymbol{y_{1:t}}, \boldsymbol{x_{1:t}};\phi)
    \end{split}
\end{equation}
}

In the inference phase, we use the basin's current driver and response data to create the internal state $z$. We then use the internal state, along with the forecasted driver data, to make predictions.

\subsection{Mitigating lack of Training Data:}
Due to the heterogeneity in environmental systems behavior, PBMs are often calibrated for each entity individually. High-quality and plentiful observations are available only for a limited number of entities, and other entities may have limited or no data. For example, amongst the lakes being studied by the United States Geological Survey (USGS), less than 1\% of the lakes have 100 or more days of temperature observations, and less than 5\% of the lakes have ten or more days of temperature observations~\cite{read2017water}. RNN-based models trained with limited observed data can lead to poor performance. Thus, this section presents two strategies for training our FHNN framework that mitigates the lack of training data.

\subsubsection{Training a single Global Model}
ML offers excellent potential for dealing with the high degree of heterogeneity that is always present in environmental systems by leveraging a collection of observations from a diverse set of entities to build a powerful global model. The reason is that ML models can benefit from training data from diverse entities and thus can transfer knowledge across entities. Previously, multi-basin models have shown improved prediction capabilities~\cite{li2022regionalization,kratzert2019towards} and have reduced input data needs~\cite{ghosh2022robust, ghosh2023entity,renganathan2025task}.

Thus, we further develop our KGML framework, FHNN$_{global}$, to jointly leverage information from multiple basins. Similar to ~\cite{li2022regionalization}, we assign a one-hot vector to each basin, with a vector of dimension equal to the number of basins. The joint optimization objective of FHNN$_{global}$ during training can be formally expressed as:

{

\begin{equation}
    \label{eq:globalLoss}
    \begin{split}
        \arg \min_{\phi, \theta} &\quad\frac{1}{NK}\sum_{i=1}^{N}\sum_{t=T+1}^{T+K} (y_i^t - \mathcal{F}(\boldsymbol{z_i}, \boldsymbol{x_i^{T+1:t}};\theta))^2\\
        where &\quad z_i = \mathcal{E}(\boldsymbol{y^{1:t}_i}, \boldsymbol{x^{1:t}_i};\phi)
    \end{split}
\end{equation}
}

\subsubsection{Pretraining with Simulation Data:}
Poor initialization can cause ML models to anchor in a local minimum, especially for deep neural networks. Further, ML models can only learn (however complex) patterns in the data used for training and thus fail on unseen data that is outside the range seen in training. On the other hand, PBMs can make predictions for any arbitrary input variables (e.g., heretofore unseen weather patterns that may result from changing climate). Hence, incorporating scientific knowledge in the ML framework can improve generalization in unseen scenarios. Additionally, if physical knowledge can help inform the initialization of the weights, model training can be accelerated, requiring fewer epochs for training and fewer training samples to achieve good performance.

To address these issues, we pre-train the FHNN model using the simulated streamflow produced by a generic SAC-SMA model, which has been lightly calibrated. These simulated streamflow data are often imperfect, but they provide a synthetic realization of the physical responses of a basin to a given set of meteorological drivers. Hence, pre-training a neural network using simulations from the SAC-SMA model allows the network to emulate synthetic but physically realistic phenomena. The model training follows a similar procedure to the standard FHNN training by just replacing the observed output ($y$) with the simulated output ($y_{sim}$), as shown:

{

\begin{equation}
    \label{eq:simLoss}
    \begin{split}
        \arg \min_{\phi, \theta} &\frac{1}{K}\sum_{t=T+1}^{T+K} (y_{sim}^t - \mathcal{F}(\boldsymbol{z}, \boldsymbol{x_{T+1:t}};\theta))^2\\
        where &\quad z = \mathcal{E}(\boldsymbol{y_{sim}^{1:t}}, \boldsymbol{x_{1:t}};\phi)
    \end{split}
\end{equation}
}

When applying the pre-trained model, we fine-tune it using true observations, as shown in Eq.~\ref{eq:Loss}. Since the pre-trained model is closer to the optimal solution, it requires fewer observations to train effectively. Our experiments demonstrate that the pre-trained models achieve high accuracy with minimal data.

\section{Experiments}
\label{sec:Experiments}

\subsection{Datasets}
We evaluate our method to predict streamflow for several river catchments from the National Weather Service (NWS) North Central River Forecast Center (NCRFC) region. Input driver data is limited to catchment-scale Mean Areal Precipitation (MAP) and Mean Areal Temperature (MAT) values at a six-hour time-step, just as needed for the NCRFC operational Sacramento Soil Moisture Accounting Model and Snow-17 suite of models. We select multiple validation basins in the NCRFC area. We divide the observed data into training and testing periods. The training period extends through 2011, with data up to 2009 used for model training and 2010-2011 data reserved for validation. We then assess model performance during the testing period, which generally spans from 2012 to 2019.

\subsection{Model Setup and Baselines}
Here, we compare model performance to multiple baselines\\
\noindent\textbf{NWS:} The National Weather Service (NWS) model is a PBM that simulates rainfall-runoff. It transforms meteorological data into comprehensive hydrological information like evapotranspiration, runoff, infiltration, groundwater, and catchment streamflow.\\
\noindent\textbf{LSTM:} We compare FHNN to a standard Long Short-Term Memory (LSTM) model that generates streamflow forecasts based on precipitation and temperature inputs.\\
\noindent\textbf{LSTM-AR:} We train an autoregressive LSTM Machine Learning model that takes in precipitation, temperature, and the most recent observed streamflow as inputs and directly generates a streamflow forecast. This autoregressive ML approach uses the same input information as our KGML approach.\\
\noindent\textbf{RR-Former~\cite{yin2022rr}:} A Transformer-based rainfall-runoff model that uses attention mechanisms to predict water runoff patterns across multiple basins.\\
\noindent\textbf{TiDE~\cite{das2023long}:} TiDE is a fast, MLP-based encoder-decoder model for long-term time series forecasting that achieves competitive performance to Transformer-based alternatives.\\
\noindent\textbf{ExoTST~\cite{tayal2024exotst}:} ExoTST is a transformer-based framework that improves time series prediction by effectively integrating past endogenous variables with both past and current exogenous variables through a novel cross-temporal modality fusion module.\\
\noindent\textbf{TFT~\cite{lim2021temporal}:} Temporal Fusion Transformer (TFT) is an interpretable attention-based architecture that combines recurrent and self-attention layers to process multiple types of inputs for high-performance multi-horizon forecasting. TFT is an Encoder-Decoder model that achieves implicit factorization due to attention over time-steps.\\

Note that the NWS and LSTM models are comparable and have equal information. Similarly, LSTM-AR, RR-Former, TiDE, ExoTST, TFT, and FHNN are comparable because all three have previous outputs available in a data assimilation setting.

\begin{table*}[!t]
    \caption{\footnotesize \textit{Model Performance Comparison (Nash-Sutcliffe Efficiency, NSE) of Hydrologic Models Across Multiple Stations}}
    \label{tab:local}
    \small
    \resizebox{\linewidth}{!}{%
        \begin{tabular}{|c|c|cc|cccccccccccccc|}
            \hline
            \textbf{Model Type}                                                     & \textbf{Model}        & \rotatebox{90}{\textbf{Mean}} & \rotatebox{90}{\textbf{Median}} & \rotatebox{90}{\textbf{AGYM5}} & \rotatebox{90}{\textbf{AMEI4}} & \rotatebox{90}{\textbf{BCHW3}} & \rotatebox{90}{\textbf{BEAW3}} & \rotatebox{90}{\textbf{BIFM5}} & \rotatebox{90}{\textbf{HWYM5}} & \rotatebox{90}{\textbf{CTVI2}} & \rotatebox{90}{\textbf{DARW3}} & \rotatebox{90}{\textbf{EAGM4}} & \rotatebox{90}{\textbf{IRNM7}} & \rotatebox{90}{\textbf{KALI4}} & \rotatebox{90}{\textbf{MRPM4}} & \rotatebox{90}{\textbf{RUSI2}} & \rotatebox{90}{\textbf{STRM4}} \\
            \hline
            \textbf{PBM}                                                            & \textbf{NWS Sac-SMA}  & 0.53          & 0.61            & 0.60           & 0.73           & 0.54           & 0.20           & 0.16           & 0.13           & 0.81           & 0.59           & 0.24           & 0.64           & 0.75           & 0.63           & 0.75           & 0.72           \\
            \hdashline
            \multirow{4}{*}{\textbf{ML}}                                            & \textbf{LSTM}         & 0.62          & 0.68            & 0.14           & 0.66           & 0.44           & 0.55           & 0.70           & 0.44           & 0.85           & 0.40           & 0.80           & 0.64           & 0.71           & 0.72           & 0.76           & 0.81           \\
                                                                                    & \textbf{LSTM-AR}      & 0.73          & 0.76            & 0.48           & 0.74           & 0.55           & 0.68           & 0.80           & 0.76           & 0.91           & 0.49           & 0.87           & 0.66           & 0.76           & 0.78           & 0.86           & 0.83           \\
                                                                                    & \textbf{RR-Former}    & 0.65          & 0.73            & 0.17           & 0.42           & 0.47           & 0.71           & 0.76           & 0.77           & 0.87           & 0.46           & 0.76           & 0.60           & 0.71           & 0.80           & 0.83           & 0.78           \\
                                                                                    & \textbf{TiDE}         & 0.45          & 0.48            & 0.51           & 0.25           & 0.22           & 0.64           & 0.69           & 0.78           & 0.40           & 0.27           & 0.71           & 0.00           & 0.22           & 0.58           & 0.55           & 0.44           \\
            \hdashline
            \multirow{1}{*}{\textbf{ML, Enc-Dec}}                                  & \textbf{ExoTST}       & 0.65          & 0.72            & 0.48           & 0.48           
                                                                                    &  0.44           & 0.72           & 0.77           & 0.78           & 0.78           & 0.47           & 0.81           & 0.49           & 0.59           & 0.79           & 0.79           & 0.72           \\
            \hdashline
            \textbf{ML, Enc-Dec, Implicit Fact$^\text{n}$}                           & \textbf{TFT}          & 0.74          & 0.77            & 0.49           & 0.70           & 0.55           & 0.75           & 0.76           & 0.79           & 0.91           & 0.50           & 0.88           & 0.67           & 0.77           & 0.81           & 0.88           & 0.85           \\
            \hdashline
            \multirow{2}{*}{\textbf{ML, Enc-Dec, Explicit Fact$^\text{n}$ }} & \textbf{FHNN$_{single}$} & 0.76          & 0.80            & 0.61           & 0.77           & 0.55           & 0.75           & 0.82           & 0.83           & \textbf{0.92}  & 0.53           & 0.87           & 0.64           & 0.78           & 0.81           & \textbf{0.90}  & 0.84           \\
                                                                                    & \textbf{FHNN}         & \textbf{0.79} & \textbf{0.83}   & \textbf{0.64}  & \textbf{0.80}  & \textbf{0.58}  & \textbf{0.77}  & \textbf{0.84}  & \textbf{0.85}  & \textbf{0.92}  & \textbf{0.60}  & \textbf{0.89}  & \textbf{0.69}  & \textbf{0.82}  & \textbf{0.83}  & \textbf{0.90}  & \textbf{0.86}  \\
            \hline
        \end{tabular}
    }
\end{table*}


\subsection{Hyperparameters and Architectures}
We create sliding windows of 748-time steps for the streamflow dataset, with a forecast length of 28 steps. Here, LSTM takes 720-length sequences (6 months of data) as input and generates output of length 28 steps (7 days). We create input sequences of length 748 timesteps using a stride of 1. All LSTMs used in the response predictor for FHNN (decoder) and the baselines have one hidden layer with 32 units, whereas the LSTMs used in the encoder of FHNN have a hidden layer with 11 units. The feed-forward network used to get the latent state has one hidden layer with 32 units. Similarly, for transformer-based baselines, we use 32 units as the dimension of feed-forward layers. Note that all models (FHNN and baselines) have a similar total number of parameters. In our experiments, we perform extensive hyperparameter search with the list provided in Appendix~\ref{sec:Hyperparameter_NWS-NCRFC}. To reduce the randomness typically expected with network initialization, we report the result of ensemble prediction obtained by averaging predictions from five models with different weight initializations for all architectures (ours and baselines).

We used the Nash-Sutcliffe Model Efficiency (NSE), a widely used metric to measure the predictive skill of hydrologic models. Equation~\ref{eq:NSE} shows the NSE equals one minus the ratio of error variance (MSE) and variance of the observations. When the error variance is larger than the variance of the observations, NSE becomes negative. As a result, NSE ranges from $(-\infty$, 1], with 1 being a perfect forecast and a negative number being poorer performance in prediction than simply offering the mean historical observation as the forecast. As a measurement metric, the higher NSE means better prediction skills for the investigated models. The NSE was calculated as an average of the NSE for all 7-day forecasts generated by the KGML model, compared to the overall NSE for the NWS model.

{
\begin{equation}
    \label{eq:NSE}
    \text{NSE}(Y, \hat{Y}) = 1 - \frac{\sum_{i=1}^{N}(\hat{Y_i} - Y_i)^2}{\sum_{i=1}^{N}(Y_i - \bar{Y})^2}
\end{equation}
\vspace{-0.1in}
}

\section{Results}
\label{sec:Results}

\subsection{Overall Predictive Performance} 
First, we aim to evaluate how creating states at different levels helps improve forecast accuracy. Table~\ref{tab:local} presents a comprehensive performance comparison across multiple NOAA sites using Nash-Sutcliffe Efficiency (NSE). The physics-based NWS Sac-SMA model, which simulates rainfall-runoff processes through comprehensive hydrological components, achieves a mean NSE of 0.53. In the ML category, the standard LSTM, operating solely on precipitation and temperature inputs, achieves a mean NSE of 0.62, while its autoregressive variant (LSTM-AR), which additionally incorporates recent streamflow observations, shows marked improvement with 0.73. The linear MLP-based model TiDE shows limited efficacy with a mean NSE of 0.45. Among Transformer-based approaches, RR-Former, designed specifically for rainfall-runoff modeling, performs reasonably well with 0.65. The transformer-based encoder-decoder architectures show promising results: ExoTST, with its cross-temporal modality fusion, achieves 0.65, while TFT, leveraging implicit factorization through its attention mechanisms, reaches 0.74. The most significant improvements come from our proposed \textit{explicit factorization} approach. FHNN$_{single}$ which involves a single factorization of hidden state, achieves an 0.76 mean NSE. The full FHNN model achieves the stat-of-the-art performance with 0.79 mean NSE, representing a 49\% improvement over the Sac-SMA baseline and a 6\% improvement over TFT. This validates our hypothesis that explicit temporal factorization better captures the multi-scale dynamics inherent in hydrological systems.

\begin{figure}
    \centering
    \includegraphics[width=\linewidth]{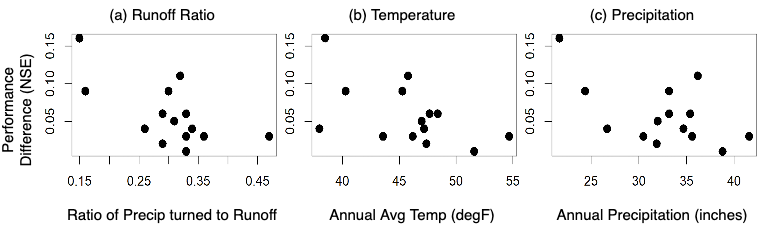}
    \caption{\footnotesize \textit{Performance difference between FHNN and LSTM-AR models plotted against (a) Runoff Ratio, (b) Annual Average Temperature, and (c) Annual Precipitation across different catchments.}}
    \label{fig:scatterPlot}
    \vspace{-0.2in}
\end{figure}

FHNN outperforms LSTM-AR in all NWS catchments tested. However, the highest performance differential is for basins that have the lowest Runoff Ratios - or, the annual proportion of incoming precipitation that is discharged from the catchment outlet as streamflow. The scatter plot shown in Figure~\ref{fig:scatterPlot} (a) reveals a strong negative correlation between runoff ratio and FHNN's performance advantage over LSTM-AR. The performance difference between FHNN and LSTM-AR is plotted against each basin's runoff ratio, showing that FHNN's greatest advantages occur in basins with low runoff ratios. A higher precipitation to runoff ratio means that annually, more of the precipitation transforms into streamflow runoff, which is easier to model (and thus FHNN has only a modest improvement over LSTM-AR). A low value means that over the course of the year, incoming water is retained on the landscape and used by plants, percolated into deeper storage, or stored and sublimated in winter snowpack to be melted later in the spring season, etc. Thus, a longer memory and better representation of catchment states is needed within the model to adequately capture the streamflow response of these catchments. This is where FHNN has the greatest improvement over LSTM-AR. For example, AGYM5 (which shows the highest NSE improvement of 0.16)  has the lowest runoff ratio (0.15). Conversely, the basins that have the least improvement in performance gains (e.g., STRM4, CTVI2) have high rainfall-runoff ratios (0.47, 0.33 respectively) the performance gains are more modest (less than 0.03 NSE). We observed similar trends in Figure~\ref{fig:scatterPlot} (b, c): where FHNN performs relatively better for colder (b) and drier (c) catchments. 
This trend is explainable because colder catchments would need additional state dynamics to represent snow accumulation, evaporation, and melt on top of rainfall-runoff processes. Drier catchments will need a longer effective memory to accurately forecast the rainfall-runoff response because runoff-inducing events will generally be rare. Drier catchments also tend to have a more strongly nonlinear rainfall-runoff response \cite{ye1997performance}.

\begin{table}[]
    \caption{\footnotesize \textit{Model Performance Comparison (Nash-Sutcliffe Efficiency, NSE) of Hydrologic Models Across Multiple Stations in Limited Data Scenario.}}
    \label{tab:local-limited}
    \resizebox{\linewidth}{!}{%
        \begin{tabular}{|c|cc|cc|cc|}
            \hline
            \multirow{2}{*}{\textbf{Model}} & \multicolumn{2}{c|}{\textbf{2 Years}} & \multicolumn{2}{c|}{\textbf{5 Years}} & \multicolumn{2}{c|}{\textbf{9+ Years}} \\
            \cline{2-7}
                                   & \textbf{Mean}         & \textbf{Median}       & \textbf{Mean}        & \textbf{Median}        & \textbf{Mean}         & \textbf{Median}        \\
            \hline
            \textbf{LSTM-AR}                & 0.56         & 0.56         & 0.67        & 0.69          & 0.73         & 0.76          \\
            \textbf{RR-Former}              & -0.19        & -0.02        & 0.22        & 0.18          & 0.65         & 0.73          \\
            \textbf{ExoTST}                 & 0.34         & 0.36         & 0.53        & 0.61          & 0.65         & 0.72          \\
            \textbf{TFT}                    & 0.45         & 0.46         & 0.61        & 0.63          & 0.74         & 0.77          \\
            \textbf{FHNN}                   & 0.58         & 0.68         & 0.70        & 0.75          & 0.79         & 0.83          \\
            \hline
        \end{tabular}
    }
    \vspace{-0.25in}
\end{table}

In investigating model performance across different deep learning architectures for hydrological prediction, our findings revealed distinct patterns related to data availability and architectural capabilities in handling path-dependent systems. The results particularly highlight the challenges faced by transformer-based models in limited data scenarios and the advantages of architectures that can effectively capture multi-scale temporal dependencies and process information hierarchically. Key differences emerged in how various architectures handled the path-dependent nature of hydrological systems, especially when training data was constrained. Models that incorporated explicit sequential processing mechanisms or physics-informed components demonstrated clear advantages in such scenarios. This aligns with the theoretical understanding that hydrological systems require tracking the evolution of various states (such as soil moisture and groundwater levels) that influence current streamflow predictions. The limited availability of training data probably compounds these challenges. Hydrological records often span less than a few decades for most locations, while material fatigue studies require extensive testing periods. These constraints may favor LSTM's inductive bias, whose architecture explicitly models sequential state evolution through internal memory states.

These theoretical expectations are strongly supported by our empirical findings summarized in Table~\ref{tab:local-limited}. With just 2 years of training data, FHNN achieved the highest performance (mean NSE = 0.58, median NSE = 0.68). In contrast, the pure transformer-based approach (RR-Former) struggled significantly, showing negative NSE values (mean NSE = -0.19, median NSE = -0.02). The performance gap between architectures narrowed as training data increased to 5 years,with the same relative ranking. FHNN maintained superior performance (mean NSE = 0.70, median NSE = 0.75), while LSTM-AR (mean NSE = 0.67, median NSE = 0.69) continued to outperform transformer-based approaches. Only with extensive training data (9+ years) did transformer-based models approach the performance levels of sequential architectures. Under these conditions, while FHNN still led (mean NSE = 0.79, median NSE = 0.83), the TFT model (mean NSE = 0.74, median NSE = 0.77) achieved comparable results to LSTM-AR (mean NSE = 0.73, median NSE = 0.76).

These results demonstrate that while transformer-based architectures can eventually learn to model path-dependent systems, they require substantially more training data to achieve performance comparable to architectures that explicitly model sequential dependencies. FHNN's consistent superior performance across all data availability scenarios underscores the importance of incorporating domain-specific biases for model architecture. Its hierarchical organization of temporal information processing, combined with its ability to handle multi-scale dynamics, proves advantageous for modeling path-dependent hydrological systems, especially in data-limited scenarios common in environmental applications.

\subsection{Pretraining with Simulation Data}
Here, we show the power of pre-training to improve the model's prediction accuracy even with small amounts of training data. A basic premise of pre-training our models is that NWS SAC-SMA simulations, though imperfect, provide a synthetic realization of physical responses of a catchment to a given set of meteorological drivers. Hence, pre-training a neural network using SAC-SMA simulations allows the network to emulate a synthetic realization of physical phenomena. We hypothesize that such a pre-trained model requires fewer labeled samples to achieve good generalization performance, even if the SAC-SMA simulations do not match the observations. To test this hypothesis, we pre-train FHNN on the simulated data by minimizing the loss function defined in Equation~\ref{eq:simLoss}. We further fine-tune the pre-trained FHNN models with two years of observed data and report the performance in Table~\ref{tab:simulation_stats}. This model is denoted using the notation ``\textbf{Sim}: Yes \textbf{Obs}: 2yr'' in Table~\ref{tab:simulation_stats}. For comparison, we also report the performance of FHNN models trained from scratch using only two years of observed data, denoted by ``\textbf{Sim}: No \textbf{Obs}: 2yr''. Finally, we report the FHNN model that has been trained using all the available data denoted by ``\textbf{Sim}: No \textbf{Obs}: All''. Table~\ref{tab:simulation_stats} shows that pre-training can significantly improve the performance. The improvement is relatively much larger, given a small amount of observed data. Even with two years of observed data, a pre-trained FHNN achieves an NSE similar to that obtained by the model when using all of the observed data.

\begin{table}[t!]
    \centering
    \caption{\footnotesize \textit{Performance comparison (mean and median NSE) of the model on increasing amount of training observation and the availability of simulated data for pretraining. The models are evaluated based on varying amount of training observations, ranging from 0.5 years to 10 years, highlighting the improvement in performance metrics with increased observation durations and the inclusion of simulations. Note that he simulations are obtained from NWS Sac-SMA model with mean and median NSE of 0.53 and 0.61 respectively.}}
    \vspace{-0.1in}
    \label{tab:simulation_stats}
    \resizebox{\columnwidth}{!}{%
        \begin{tabular}{|c|cc|cc|}
            \hline
            \multirow{2}{*}{\textbf{Finetuning Data}} & \multicolumn{2}{c|}{\textbf{Sim: No}} & \multicolumn{2}{c|}{\textbf{Sim: Yes}} \\
            \cline{2-5}
                                             & \textbf{Mean}    & \textbf{Median}   & \textbf{Mean}    & \textbf{Median}    \\
            \hline
            \textbf{Obs: 0.5 yr}             & 0.18             & 0.30              & 0.68             & 0.75               \\
            \textbf{Obs: 1 yr}               & 0.41             & 0.42              & 0.74             & 0.77               \\
            \textbf{Obs: 2 yr}               & 0.58             & 0.68              & 0.77             & 0.81               \\
            \textbf{Obs: 10 yr}              & 0.79             & 0.83              & 0.78             & 0.82               \\
            \hline
        \end{tabular}%
    }
    \vspace{-0.2in}
\end{table}

Thus, FHNN has shown that it can learn from a physically-based model and only a small amount of observed data, thereby acting as a model assimilator. These insights from process-based modeling can be used in regions with limited streamflow data. The service ramifications of this result are clear – often, communities have particular stream reaches of concern that cause localized flooding and desire RFC and NWS services to produce forecasts. However, funding is also often limited, and these communities perhaps fund a stream gage for a year or two, hoping it will be enough for NWS to begin forecast services. Calibrated process-based models, however, often need at least five years of data to ensure robust calibration for NWS RFC forecast services. The ability of the KGML to learn from a minimal amount of data with the help of model globalization and the use of imperfect process-based models means that potentially years will be taken off the time required to develop robust RFC forecast services. Although additional verification of model robustness in different conditions is underway, initial results show that this is a vision for the not-too-distant future.

\begin{table}[]
    \centering
    \caption{\footnotesize \textit{Performance comparison (median NSE) of the models to their global counterparts. The models are evaluated based on varying amount of training observations, ranging from 1 year to 10 years.}}
    \vspace{-0.1in}
    \label{tab:global}
    \resizebox{0.8\columnwidth}{!}{%
        \begin{tabular}{|c|ccc|ccc|}
            \hline
            \multirow{2}{*}{\textbf{Model}} & \multicolumn{3}{c|}{\textbf{Local}}                          & \multicolumn{3}{c|}{\textbf{Global}}                         \\ \cline{2-7} 
             &
              \multicolumn{1}{c|}{\textbf{1 yr}} &
              \multicolumn{1}{c|}{\textbf{2 yr}} &
              \textbf{10 yr} &
              \multicolumn{1}{c|}{\textbf{1 yr}} &
              \multicolumn{1}{c|}{\textbf{2 yr}} &
              \textbf{10 yr} \\ \hline
            \textbf{LSTM-AR}                & \multicolumn{1}{c|}{0.30} & \multicolumn{1}{c|}{0.56} & 0.68 & \multicolumn{1}{c|}{0.34} & \multicolumn{1}{c|}{0.61} & 0.77 \\
            \textbf{FHNN} &
              \multicolumn{1}{c|}{\textbf{0.42}} &
              \multicolumn{1}{c|}{\textbf{0.68}} &
              \textbf{0.83} &
              \multicolumn{1}{c|}{\textbf{0.49}} &
              \multicolumn{1}{c|}{\textbf{0.74}} &
              \textbf{0.84} \\ \hline
        \end{tabular}%
    }
    \vspace{-0.25in}
\end{table}

\subsection{Learning Global Model}
Table \ref{tab:global} presents a performance comparison between local and global versions of LSTM-AR and FHNN models, evaluated using median NSE scores across varying amounts of training data. The results show that FHNN consistently outperforms LSTM-AR across all training durations in local and global settings. For local models, FHNN$_{local}$ achieves higher median NSE scores of 0.42, 0.67, and 0.78 for 1-year, 2-year, and 10-year training periods, respectively, compared to LSTM-AR$_{local}$'s scores of 0.30, 0.56, and 0.68. The global models demonstrate improved performance over their local counterparts, with FHNN$_{global}$ again leading with scores of 0.49, 0.70, and 0.80 for the same training periods. Notably, the performance gap between FHNN$_{global}$ and LSTM-AR$_{global}$ is more pronounced in scenarios with limited training data (1-year), highlighting FHNN$_{global}$ effectiveness in learning from smaller datasets. Both models show significant improvement as the training data increases to 10 years, but FHNN$_{global}$ maintains its superior performance. Thus, the table clearly illustrates that the global configurations of these models generally lead to improved NSE performance compared to their standard counterparts.

\vspace{-0.1in}
\subsection{Visualizing States}
The State Encoder ($\mathcal{E}$) forms a key component of our proposed FHNN as it allows for a deeper understanding and interpretation of the complex dynamics governing physical systems. The inferred states encapsulate the system's inherent characteristics and behaviors, dictating how the system responds to external influences. FHNN builds states at different temporal scales, capturing both short-term fluctuations and long-term trends. This multi-scale state construction enhances our ability to model and predict the behavior of diverse physical systems, providing a comprehensive understanding and unlocking new insights into their complex dynamics. Thus, it is necessary to evaluate the quality of these inferred states.

\begin{figure}[t!]
    \centering
    \includegraphics[width=\columnwidth]{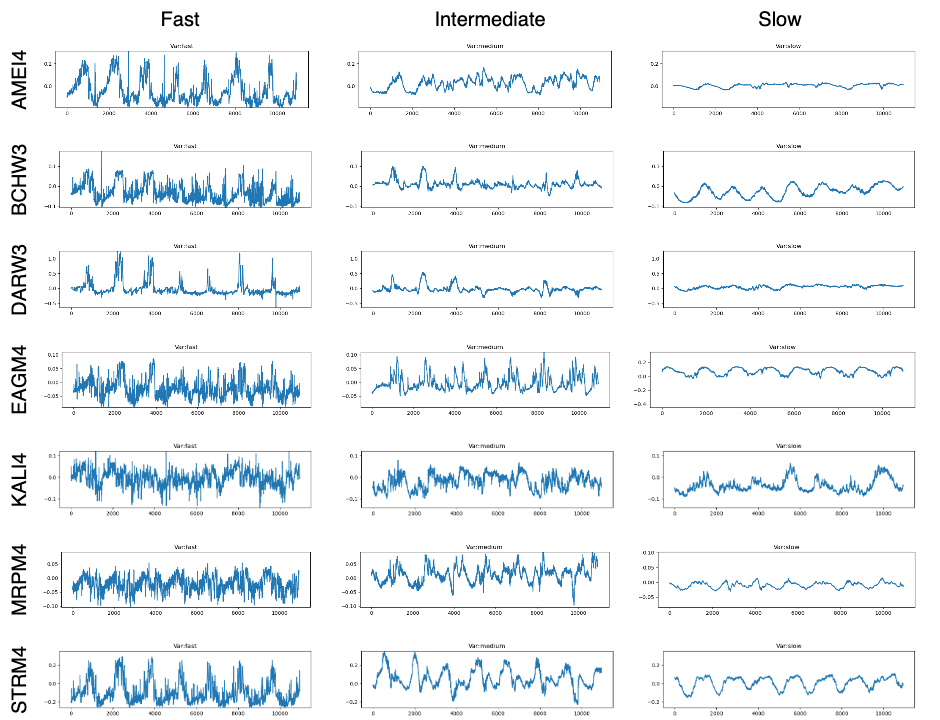}
    \vspace{-0.2in}
    \caption{\footnotesize \textit{Visualization of FHNN's inferred fast, medium, and slow states for seven river basins. Each row represents a basin, with columns showing the mean of hidden state dimensions over time for each temporal scale. Fast states (left) exhibit high variability, medium states (center) show smoothed patterns, and slow states (right) capture long-term trends.}}
    \label{fig:States}
    \vspace{-0.25in}
\end{figure}

In Figure~\ref{fig:States}, we visualize the slow, medium, and fast states from the model for AMEI4, BCHW3, DARW3, EAGM4, KAL4, MRPM4, and STRM4 basins. Recall that each LSTM (slow, medium, and fast) in the state encoder has a hidden state dimension of eleven. In this figure we take the mean of the dimensions for each corresponding state and plot them (y-axis) w.r.t. to time (x-axis). This visualization allows for the comparison of different temporal scales of variation within and between the variables, from rapid fluctuations to long-term trends. We plot the fast, medium, and slow states for each basin arranged in columns. The first column with ``fast" states exhibits the most variability and detail. The middle column with the ``medium" states appears to be a smoothed/filtered version with less high-frequency variation. The last column with ``slow" states shows an even more smoothed version, capturing only the slowest/long-term trends that are not immediately apparent in the raw data.

\subsection{Evaluation on CAMELS}
FHNN is evaluated on CAMELS (Catchment Attributes and MEteorology for Large-sample Studies), a widely used benchmarking dataset for streamflow prediction~\cite{addor2017camels}. CAMELS compiles daily meteorological forcing data (e.g., precipitation, air temperature), streamflow observation, calibrated physical model simulation, and catchment characteristics(see Appendix~\ref{sec:camels_dataset} for a complete list). Our study uses data from 531 basins from CAMELS for the period (1989-2009). The period (1989-2001) is used for training, and (2001-2009) for testing. We create input sequences of length 365 days using a stride of 1 day. Here, LSTM takes 365-length sequences as input and generates output at a length of 1, 3, and 7 days. All LSTMs used in the response predictor for FHNN (decoder) and the baselines have one hidden layer with 255 units, whereas the LSTMs used in the encoder of FHNN have a hidden layer with 85 units. The feed-forward network used to get the latent state also has one hidden layer with 255 units. All models were trained as global models (i.e., one model for all basins). We perform an extensive hyperparameter search with the list in Appendix~\ref{sec:Hyperparameter_CAMELS}. To reduce the randomness typically expected with network initialization, we train five models with different initialization of deep learning model weights. The predictions were then further combined into an ensemble by averaging predictions from these five models.

\begin{table}[t!]
    \caption{\footnotesize \textit{Performance comparison (mean and median NSE) of the model for different forecasting steps (denoted in columns) on the CAMELS dataset.}}
    \vspace{-0.1in}
    \label{tab:camels_results}
    \resizebox{\columnwidth}{!}{%
        \begin{tabular}{|c|cc|cc|cc|}
            \hline
            \multirow{2}{*}{\textbf{Model}} & \multicolumn{2}{c|}{\textbf{1 Steps}} & \multicolumn{2}{c|}{\textbf{3 Steps}} & \multicolumn{2}{c|}{\textbf{7 Steps}} \\
            \cline{2-7} 
                                            & \textbf{Mean}    & \textbf{Median}   & \textbf{Mean}    & \textbf{Median}   & \textbf{Mean}    & \textbf{Median}   \\
            \hline
            \textbf{LSTM AR}                & 0.66             & 0.80              & 0.65             & 0.79              & 0.63             & 0.78              \\
            \textbf{RR-FORMER}              & 0.65             & 0.78              & 0.59             & 0.75              & 0.53             & 0.72              \\
            \textbf{ExoTST}                 & 0.73             & 0.79              & 0.67             & 0.75              & 0.64             & 0.74                  \\
            \textbf{TFT}                    & 0.75             & 0.83              & 0.72             & 0.81              & 0.71             & 0.80                  \\
            \textbf{FHNN}                   & 0.77             & 0.85              & 0.73             & 0.82              & 0.72             & 0.81              \\
            \hline
        \end{tabular}
    }
    \vspace{-0.25in}
\end{table}

The performance comparison across various forecasting steps is presented in Table \ref{tab:camels_results}, showing mean and median NSE values for one-day, three-day, and seven-day forecasts. Our results show that FHNN achieves the highest median NSE scores across catchments (0.85, 0.82, and 0.81 for days 1, 3, and 7, respectively). As the forecast step increases, all models show a slight decrease in performance, but FHNN still maintains its superiority. This superior performance can be attributed to FHNN's hierarchically factorized architecture, which builds multi-scale states capturing both short-term dynamics and long-term patterns in streamflow behavior.
While TFT shows slightly worse mean NSE values (0.75 vs 0.77 at day 1), FHNN's higher median and mean values demonstrate reliable performance across the full spectrum of catchment behaviors. This reliability stems from its hierarchical state factorization, which explicitly models the multi-scale temporal dependencies inherent in hydrological systems.
As demonstrated in Section 6.2, FHNN maintains strong performance even with limited training data, a critical advantage since most global watersheds have less than 2-3 years of historical records. While transformer-based models achieve competitive results on CAMELS' extensive 10+ years of training dataset, they struggle in data-limited scenarios. FHNN's LSTM-based hierarchical architecture provides consistent performance across both data conditions, making it uniquely suited for widespread deployment.
These results validate FHNN's design principle of modeling hydrological systems' multi-scale temporal dependencies through hierarchical state factorization, addressing key challenges in operational hydrological forecasting.

\vspace{-0.1in}
\section{Conclusion}
\label{sec:Conclusion}
This paper presents a novel approach to modeling dynamical systems using a factorized hierarchical neural network (FHNN). The proposed method demonstrates significant improvements in predictive performance under the data assimilation setting in hydrologic forecasting. The FHNN model effectively captures complex dependencies in sequential data by incorporating hierarchical state representations at multiple temporal scales. Experimental results of our study demonstrate the effectiveness of the FHNN model in improving hydrologic forecasting compared to traditional models like the NWS Sac-SMA, LSTM-based auto-regressive approaches, and transformer-based encoder-decoder approaches for streamflow forecasting using data from the National Weather Service North Central River Forecast Center. FHNN exhibits the strongest performance improvements in basins with low runoff ratios and cold climates, where capturing long-term dependencies and complex snow accumulation dynamics proves crucial. Further, FHNN can maintain performance advantages in data-limited scenarios through effective pre-training strategies and global model variants. Further, evaluations on the CAMELS benchmark dataset confirm FHNN's effectiveness across diverse hydrological conditions. While demonstrated for streamflow forecasting, FHNN's ability to model hierarchical temporal dependencies makes it applicable to other domains requiring multi-scale dynamical system modeling.

\section*{Acknowledgment}

This work was supported by NSF grant (2313174) and the NSF LEAP Science and Technology Center (award 2019625). This work was also supported by a Data Science Graduate Assistantship with funding made available by the MnDRIVE initiative through the University of Minnesota Data Science Initiative. This research is also part of AI-CLIMATE: "AI Institute for Climate-Land Interactions, Mitigation, Adaptation, Tradeoffs and Economy," funded by the USDA National Institute of Food and Agriculture (NIFA) and the National Science Foundation (NSF) National AI Research Institutes Competitive Award no. 2023-67021-39829. Computational resources were provided by the Minnesota Supercomputing Institute.


\bibliographystyle{acm}
\bibliography{main}

\newpage
\appendix
\section{NWS Model and Operations}
The rainfall-runoff transformation, influenced by various meteorological and land-surface characteristics, is modeled using a suite of physics-based models namely, NWS Snow-17~\cite{anderson1976point} model for snowmelt simulation and the Sacramento Soil Moisture Accounting Model (Sac-SMA)~\cite{burnash1973generalized,fowler2016simulating,john2021disaggregated} for rainfall-runoff transformation. The combined Snow-17 and Sac-SMA models simulate various hydrological processes and contain approximately 18 major parameters that are calibrated manually~\cite{anderson2002calibration}. Daily manual assimilation of observed streamflows into the calibrated PBMs by altering their underlying states is a crucial component of NWS RFC operations, as the PBMs themselves are not automatically updated by observations. While necessary, this manual assimilation is time-consuming and introduces uncertainty, highlighting the need for more efficient and accurate methods of integrating real-time data into hydrological forecasting models. Figure~\ref{fig:related work} (a) shows the diagrammatic representation of the PBM.

\section{CAMELS Dataset}
\label{sec:camels_dataset}
The Catchment Attributes and Meteorology for Large-sample Studies (CAMELS) dataset \cite{addor2017camels} is a comprehensive hydrological benchmark resource curated by the National Center for Atmospheric Research (NCAR) for the continental United States. It encompasses data for 531 basins, providing daily basin-averaged Maurer forcings \cite{maurer2002long} for time-dependent meteorological inputs, which include five key variables (as shown in Table \ref{tab:met_variables_camels}), alongside 27 static catchment characteristics (detailed in Table \ref{tab:met_variables_camels}).

\begin{table}[htbp]
    \resizebox{\linewidth}{!}{
        \begin{tabular}{|l|l|l|}
        \hline
        \textbf{Meteorological forcing data}                                               &\textbf{Description} & \textbf{Unit} \\ \hline
        PRCP                                 &Daily Precipitation sum& $mm/day$        \\ \hline %
        SRAD                    &Mean Shortwave radiation& $W/m^2$          \\ \hline %
        Tmax                                      &Maximum Air temperature& $^\circ C$            \\ \hline %
        Tmin                                      &Minimum Air temperature& $^\circ C$            \\ \hline %
        Vp                                    &Vapor pressure& $Pa$           \\ \hline %
        \end{tabular}
    }
    \caption{\small Summary of CAMELS's meteorological variables.}
    \label{tab:met_variables_camels}
\end{table}

\begin{table}[htbp]
    \small
    \centering
    \begin{tabular}{|c|c|}
        \hline
        Group                  & Name\\
        \hline
        Climate               & p mean \\       
                            & pet mean \\  
	                         & p seasonality\\	    
	                      & frac snow\\	     
	                  & aridity\\	  
	                 & high prec freq\\	   
	                  & high prec dur\\	   
                    	& low prec freq\\
                	& low prec dur\\
        \hline
        Soil geology	      & carbonate rocks frac\\
                           	& geol permeability\\
                         	& soil depth pelletier\\
                           	& soil depth statsgo\\
                        	& soil porosity	\\
                           	& soil conductivity\\
                           	& max water content\\
                             	& sand frac	\\
                           	& silt frac	\\
                           	& clay frac\\
        \hline
        Geomorphology	      & elev mean\\
                           	& slope mean\\
                         	& area gages2\\
                        	& frac forest\\
                       	& lai max\\
                          	& lai diff\\
                        	& gvf max\\
                       	& gvf diff\\
        \hline
    \end{tabular}
    \caption{\small Table of static catchment characteristics used in this experiment, description of the characteristics is available and defined in \cite{addor2017camels}}
    \label{tab:camels_static}
\end{table}

\section{Hyperparamater Tuning}

\subsection{NWS-NCRFC}
\label{sec:Hyperparameter_NWS-NCRFC}
We performed a grid search across a range of parameter values to find the best hyperparameters. The possible values considered are listed in Table~\ref{tab:hyperparameter_tuning_noaa_fhnn}. We trained our model, FHNN, and other baselines using data from training years. The final hyperparameters, including batch size and learning rates, are determined after performing a k-fold cross-validation (k=5)
on the training set. Finally, we evaluated the model's performance on the held-out test period.\\

\begin{table}[htbp]
\scriptsize
\resizebox{\linewidth}{!}{
\begin{tabular}{|l|l|}
\hline
\textbf{Hyperparameter}                     & \textbf{Value}             \\ \hline
Dimension of hidden layer for Encoder                & 6, 8, \textbf{11}, 16, 32                \\ \hline
Dimension of hidden layer for Decoder & 18, 24, \textbf{33}, 48, 96               \\ \hline
Sequence Length                                 & 180,270,365,\textbf{720}             \\ \hline
Batch size                                  & 32, \textbf{64}, 128                \\ \hline
Learning rate                               & 0.005, \textbf{0.001}, 0.0005, 0.01 \\ \hline
\end{tabular}
}
\caption{\small Range of parameter values tried for hyperparameter tuning for NWS-NCRFC dataset, with the final selected value shown in \textbf{bold}.}
\vspace{-0.2cm}
\label{tab:hyperparameter_tuning_noaa_fhnn}
\end{table}

Table~\ref{tab:Hyperparameter_noaa_transformer} provides detailed information about the hyperparameter configurations for transformer-based baselines used in our experiments. 

\begin{table}[htbp]
\small
\centering
\caption{\small Hyperparameter Configuration for NOAA Dataset}
\begin{tabular}{lccc}
\toprule
\textbf{Hyperparameter} & \textbf{ExoTST} & \textbf{RRFormer} & \textbf{TFT} \\
\midrule
Model Dim & 16 & 32 & 32 \\
Forward Dim & 32 & 32 & 32 \\
Learning Rate & 1e-3 & 1e-3 & 1e-3 \\
Dropout & 0.4 & 0.4 & 0.4 \\
Number of Heads & 4 & 4 & 4 \\
Encoder Layers & 2 & 2 & 1 \\
Decoder Layers & 2 & 2 & 1 \\
Patch Size & 30 & - & - \\
Stride & 15 & - & - \\
Batch Size & 64 & 64 & 64 \\
\bottomrule
\end{tabular}
\label{tab:Hyperparameter_noaa_transformer}
\end{table}

\subsection{CAMELS}
\label{sec:Hyperparameter_CAMELS}
We performed a grid search across a range of parameter values to find the best hyperparameters. The possible values considered are listed in Table~\ref{tab:hyperparameter_tuning_camels}. We trained our model, FHNN, using data from training training. The final hyperparameters, including batch size and learning rates, are determined after performing a k-fold cross-validation (k=5)
on the training set. Finally, we evaluated the model's performance on the held-out test period.\\

\begin{table}[htbp]
\scriptsize
\resizebox{\linewidth}{!}{
\begin{tabular}{|l|l|}
\hline
\textbf{Hyperparameter}                     & \textbf{Value}             \\ \hline
Dimension of hidden layer for Encoder                & 25, 45, \textbf{85}, 128            \\ \hline
Dimension of hidden layer for Decoder & 75, 135 \textbf{255},384               \\ \hline
Sequence Length                                 & 90,180,\textbf{360}, 720             \\ \hline
Batch size                                  & 32, \textbf{64}, 128                \\ \hline
Learning rate                               & 0.005, \textbf{0.001}, 0.0005, 0.01 \\ \hline
\end{tabular}
}
\caption{\small Range of parameter values tried for hyperparameter tuning for CAMELS dataset, with the final selected value shown in \textbf{bold}.}
\vspace{-0.2cm}
\label{tab:hyperparameter_tuning_camels}
\end{table}

\begin{table}[htbp]
\small
\centering

\caption{\small Hyperparameter Configuration for CAMELS Dataset}
\begin{tabular}{lccc}
\toprule
\textbf{Hyperparameter} & \textbf{ExoTST} & \textbf{RRFormer} & \textbf{TFT} \\
\midrule
Model Dim & 96 & 96 & 128 \\
Forward Dim & 128 & 128 & 128 \\
Learning Rate & 5e-4 & 1e-4 & 1e-3 \\
Dropout & 0.4 & 0.4 & 0.4 \\
Number of Heads & 8 & 4 & 4 \\
Encoder Layers & 3 & 4 & 1 \\
Decoder Layers & 3 & 4 & 1 \\
Patch Size & 16 & - & - \\
Stride & 16 & - & - \\
Batch Size & 128 & 128 & 128 \\
\bottomrule
\end{tabular}
\label{tab:Hyperparameter_CAMELS_transformer}
\end{table}

\end{document}